\documentclass{article}

\usepackage{times}
\usepackage{graphicx} 
\usepackage{subfigure} 

\usepackage{natbib}

\usepackage{array}
\usepackage{amsmath}
\allowdisplaybreaks

\usepackage{algorithm}
\usepackage{algorithmic}

\usepackage{hyperref}

\usepackage[accepted]{icml2017} 

\icmltitlerunning{DeepLIFT: Learning Important Features Through Propagating Activation Differences}

\begin{document} 

\twocolumn[
\icmltitle{Learning Important Features Through Propagating Activation Differences}




\begin{icmlauthorlist}
\icmlauthor{Avanti Shrikumar}{stanford}
\icmlauthor{Peyton Greenside}{stanford}
\icmlauthor{Anshul Kundaje}{stanford}
\end{icmlauthorlist}

\icmlaffiliation{stanford}{Stanford University, Stanford, California, USA}

\icmlcorrespondingauthor{A Kundaje}{akundaje@stanford.edu}

\icmlkeywords{deeplift, DeepLIFT, importance scores, black box, neural networks, deep learning, visualization, understanding, propagating activation differences}

\vskip 0.3in
]



\printAffiliationsAndNotice{} 

\begin{abstract} 
The purported ``black box'' nature of neural networks is a barrier to adoption in applications where interpretability is essential. Here we present DeepLIFT (Deep Learning Important FeaTures), a method for decomposing the output prediction of a neural network on a specific input by backpropagating the contributions of all neurons in the network to every feature of the input. DeepLIFT compares the activation of each neuron to its `reference activation' and assigns contribution scores according to the difference. By optionally giving separate consideration to positive and negative contributions, DeepLIFT can also reveal dependencies which are missed by other approaches. Scores can be computed efficiently in a single backward pass. We apply DeepLIFT to models trained on MNIST and simulated genomic data, and show significant advantages over gradient-based methods. Video tutorial: \url{http://goo.gl/qKb7pL}, ICML slides: \url{bit.ly/deeplifticmlslides}, ICML talk: \url{https://vimeo.com/238275076}, code:
\url{http://goo.gl/RM8jvH}.
\end{abstract} 

\section{Introduction}
\label{introduction}

As neural networks become increasingly popular, their “black box” reputation is a barrier to adoption when interpretability is paramount. Here, we present DeepLIFT (Deep Learning Important FeaTures), a novel algorithm to assign importance score to the inputs for a given output. Our approach is unique in two regards. First, it frames the question of importance in terms of differences from a `reference' state, where the `reference' is chosen according to the problem at hand. In contrast to most gradient-based methods, using a difference-from-reference allows DeepLIFT to propagate an importance signal even in situations where the gradient is zero and avoids artifacts caused by discontinuities in the gradient. Second, by optionally giving separate consideration to the effects of positive and negative contributions at nonlinearities, DeepLIFT can reveal dependencies missed by other approaches. As DeepLIFT scores are computed using a backpropagation-like algorithm, they can be obtained efficiently in a single backward pass after a prediction has been made.

\section{Previous Work}

This section provides a review of existing approaches to assign importance scores for a given task and input example.

\subsection{Perturbation-Based Forward Propagation Approaches}

These approaches make perturbations to individual inputs or neurons and observe the impact on later neurons in the network. Zeiler \& Fergus \cite{zeiler2013visualizing} occluded different segments of an input image and visualized the change in the activations of later layers. ``In-silico mutagenesis'' \cite{zhou2015predicting} introduced virtual mutations at individual positions in a genomic sequence and quantified the their impact on the output. Zintgraf et al. \cite{zintgraf2017Visualizing} proposed a clever strategy for analyzing the difference in a prediction after marginalizing over each input patch. However, such methods can be computationally inefficient as each perturbation requires a separate forward propagation through the network. They may also underestimate the importance of features that have saturated their contribution to the output (\textbf{Fig. 1}).

\begin{figure}[htb]
\centering
\includegraphics[width=0.35\textwidth]{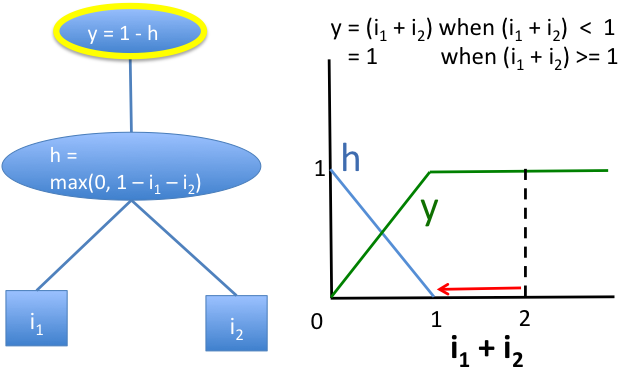}
\label{fig:saturation}
\caption{{\bf Perturbation-based approaches and gradient-based approaches fail to model saturation}. Illustrated is a simple network exhibiting saturation in the signal from its inputs. At the point where $i_1 = 1$ and $i_2=1$, perturbing either $i_1$ or $i_2$ to 0 will not produce a change in the output. Note that the gradient of the output w.r.t the inputs is also zero when $i_1 + i_2 > 1$.}
\end{figure}


\subsection{Backpropagation-Based Approaches}

Unlike perturbation methods, backpropagation approaches propagate an importance signal from an output neuron backwards through the layers to the input in one pass, making them efficient. DeepLIFT is one such approach.

\subsubsection{Gradients, Deconvolutional Networks and Guided Backpropagation}

Simonyan et al. \cite{simonyan2013deep} proposed using the gradient of the output w.r.t. pixels of an input image to compute a ``saliency map'' of the image in the context of image classification tasks. The authors showed that this was similar to deconvolutional networks \cite{zeiler2013visualizing} except for the handling of the nonlinearity at rectified linear units (ReLUs). When backpropagating importance using gradients, the gradient coming into a ReLU during the backward pass is zero'd out if the input to the ReLU during the forward pass is negative. By contrast, when backpropagating an importance signal in deconvolutional networks, the importance signal coming into a ReLU during the backward pass is zero'd out if and only if it is negative, with no regard to sign of the input to the ReLU during the forward pass. Springenberg et al., \cite{SpringenbergDBR14} combined these two approaches into Guided Backpropagation, which zero's out the importance signal at a ReLU if either the input to the ReLU during the forward pass is negative or the importance signal during the backward pass is negative. Guided Backpropagation can be thought of as equivalent to computing gradients, with the caveat that any gradients that become negative during the backward pass are discarded at ReLUs. Due to the zero-ing out of negative gradients, both guided backpropagation and deconvolutional networks can fail to highlight inputs that contribute negatively to the output. Additionally, none of the three approaches would address the saturation problem illustrated in \textbf{Fig. 1}, as the gradient of $y$ w.r.t. $h$ is negative (causing Guided Backprop and deconvolutional networks to assign zero importance), and the gradient of $h$ w.r.t both $i_1$ and $i_2$ is zero when $i_1 + i_2 > 1$ (causing both gradients and Guided Backprop to be zero). Discontinuities in the gradients can also cause undesirable artifacts (\textbf{Fig. 2}).


\subsubsection{Layerwise Relevance Propagation and Gradient $\times$ input}

Bach et al. \cite{Bach2015-pq} proposed an approach for propagating importance scores called Layerwise Relevance Propagation (LRP). Shrikumar et al. and Kindermans et al. \cite{shrikumar2016not,kindermans2016investigating} showed that absent modifications to deal with numerical stability, the LRP rules for ReLU networks were equivalent within a scaling factor to an elementwise product between the saliency maps of Simonyan et al. and the input (in other words, gradient $\times$ input). In our experiments, we compare DeepLIFT to gradient $\times$ input as the latter is easily implemented on a GPU, whereas (at the time of writing) LRP did not have a GPU implementation available to our knowledge.

While gradient $\times$ input is often preferable to gradients alone as it leverages the sign and strength of the input, it still does not address the saturation problem in \textbf{Fig. 1} or the thresholding artifact in \textbf{Fig. 2}.

\subsubsection{Integrated Gradients}

Instead of computing the gradients at only the current value of the input, one can integrate the gradients as the inputs are scaled up from some starting value (eg: all zeros) to their current value \cite{SundararajanTY16}. This addressess the saturation and thresholding problems of \textbf{Fig. 1} and \textbf{Fig. 2}, but numerically obtaining high-quality integrals adds computational overhead. Further, this approach can still give misleading results (see \textbf{Section 3.4.3}).

\subsection{Grad-CAM and Guided CAM}

Grad-CAM \cite{DBLP:journals/corr/SelvarajuDVCPB16} computes a coarse-grained feature-importance map by associating the feature maps in the final convolutional layer with particular classes based on the gradients of each class w.r.t. each feature map, and then using the weighted activations of the feature maps as an indication of which inputs are most important. To obtain more fine-grained feature importance, the authors proposed performing an elementwise product between the scores obtained from Grad-CAM and the scores obtained from Guided Backpropagation, termed Guided Grad-CAM. However, this strategy inherits the limitations of Guided Backpropagation caused by zero-ing out negative gradients during backpropagation. It is also specific to convolutional neural networks.

\section{The DeepLIFT Method}

\subsection{The DeepLIFT Philosophy}

DeepLIFT explains the difference in output from some `reference' output in terms of the difference of the input from some `reference' input. The `reference' input represents some default or `neutral' input that is chosen according to what is appropriate for the problem at hand (see \textbf{Section 3.3} for more details). Formally, let $t$ represent some target output neuron of interest and let $x_1, x_2, ..., x_n$ represent some neurons in some intermediate layer or set of layers that are necessary and sufficient to compute $t$. Let $t^0$ represent the reference activation of $t$. We define the quantity $\Delta t$ to be the difference-from-reference, that is $\Delta t = t - t^0$. DeepLIFT assigns contribution scores $C_{\Delta x_i \Delta t}$ to $\Delta x_i$ s.t.:

\begin{equation} \label{eq:1}
\sum_{i=1}^n C_{\Delta x_i \Delta t} = \Delta t
\end{equation}

We call \textbf{Eq. 1} the {\bf summation-to-delta} property. $C_{\Delta x_i \Delta t}$ can be thought of as the amount of difference-from-reference in $t$ that is attributed to or `blamed' on the difference-from-reference of $x_i$. Note that when a neuron's transfer function is well-behaved, the output is locally linear in its inputs, providing additional motivation for \textbf{Eq. 1}.

$C_{\Delta x_i \Delta t}$ can be non-zero even when $\frac{\partial t}{\partial x_i}$ is zero. This allows DeepLIFT to address a fundamental limitation of gradients because, as illustrated in \textbf{Fig. 1}, a neuron can be signaling meaningful information even in the regime where its gradient is zero. Another drawback of gradients addressed by DeepLIFT is illustrated in \textbf{Fig. 2}, where the discontinuous nature of gradients causes sudden jumps in the importance score over infinitesimal changes in the input. By contrast, the difference-from-reference is continuous, allowing DeepLIFT to avoid discontinuities caused by bias terms.

\begin{figure}[htb]
\centering
\includegraphics[width=0.35\textwidth]{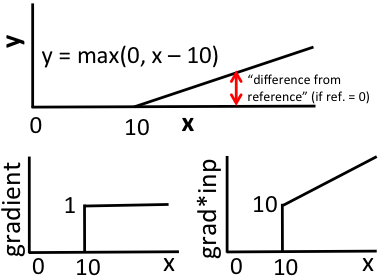}
\label{fig:saturation}
\caption{{\bf Discontinuous gradients can produce misleading importance scores}. Response of a single rectified linear unit with a bias of $-10$. Both gradient and gradient$\times$input have a discontinuity at $x=10$; at $x=10+\epsilon$, gradient$\times$input assigns a contribution of $10+\epsilon$ to $x$ and $-10$ to the bias term ($\epsilon$ is a small positive number). When $x < 10$, contributions on $x$ and the bias term are both $0$. By contrast, the difference-from-reference (red arrow, top figure) gives a continuous increase in the contribution score.}
\end{figure}

\subsection{Multipliers and the Chain Rule}

\subsubsection{Definition of Multipliers}

For a given input neuron $x$ with difference-from-reference $\Delta x$, and target neuron $t$ with difference-from-reference $\Delta t$ that we wish to compute the contribution to, we define the multiplier $m_{\Delta x \Delta t}$ as:

\begin{equation}
m_{\Delta x \Delta t} = \frac{C_{\Delta x \Delta t}}{\Delta x}
\end{equation}

In other words, the multiplier $m_{\Delta x \Delta t}$ is the contribution of $\Delta x$ to $\Delta t$ divided by $\Delta x$. Note the close analogy to the idea of partial derivatives: the partial derivative $\frac{\partial t}{\partial x}$ is the infinitesimal change in $t$ caused by an infinitesimal change in $x$, divided by the infinitesimal change in $x$. The multiplier is similar in spirit to a partial derivative, but over finite differences instead of infinitesimal ones.

\subsubsection{The Chain Rule For Multipliers}

Assume we have an input layer with neurons $x_1,..., x_n$, a hidden layer with neurons $y_1,...,y_n$, and some target output neuron $t$. Given values for $m_{\Delta x_i \Delta y_j}$ and $m_{\Delta y_j \Delta t}$, the following definition of $m_{\Delta x_i \Delta t}$ is consistent with the summation-to-delta property in \textbf{Eq. 1} (see \textbf{Appendix A} for the proof):

\begin{equation}
m_{\Delta x_i \Delta t} = \sum_j m_{\Delta x_i \Delta y_j} m_{\Delta y_j \Delta t}
\end{equation}

We refer to \textbf{Eq. 3} as the {\bf chain rule for multipliers}. Given the multipliers for each neuron to its immediate successors, we can compute the multipliers for any neuron to a given target neuron efficiently via backpropagation - analogous to how the chain rule for partial derivatives allows us to compute the gradient w.r.t. the output via backpropagation.

\subsection{Defining the Reference}

When formulating the DeepLIFT rules described in \textbf{Section 3.5}, we assume that the reference of a neuron is its activation on the reference input. Formally, say we have a neuron $y$ with inputs $x_1, x_2,...$ such that $y = f(x_1, x_2, ...)$. Given the reference activations $x_1^0, x_2^0, ...$ of the inputs, we can calculate the reference activation $y^0$ of the output as:

\begin{equation}
y^0 = f(x_1^0, x_2^0, ...)
\end{equation}

i.e. references for all neurons can be found by choosing a reference input and propagating activations through the net.

The choice of a reference input is critical for obtaining insightful results from DeepLIFT. In practice, choosing a good reference would rely on domain-specific knowledge, and in some cases it may be best to compute DeepLIFT scores against multiple different references. As a guiding principle, we can ask ourselves ``what am I interested in measuring differences against?''. For MNIST, we use a reference input of all-zeros as this is the background of the images. For the binary classification tasks on DNA sequence inputs (strings over the alphabet \{A,C,G,T\}), we obtained sensible results using either a reference input containing the expected frequencies of ACGT in the background (\textbf{Fig. 5}), or by averaging the results over multiple reference inputs for each sequence that are generated by shuffling each original sequence (\textbf{Appendix J}). For CIFAR10 data, we found that using a blurred version of the original image as the reference highlighted outlines of key objects, while an all-zeros reference highlighted hard-to-interpret pixels in the background (\textbf{Appendix L}).

It is important to note that gradient$\times$input implicitly uses a reference of all-zeros (it is equivalent to a first-order Taylor approximation of gradient$\times$$\Delta$input where $\Delta$ is measured w.r.t. an input of zeros). Similary, integrated gradients (\textbf{Section 2.2.3}) requires the user to specify a starting point for the integral, which is conceptually similar to specifying a reference for DeepLIFT. While Guided Backprop and pure gradients don't use a reference, we argue that this is a limitation as these methods only describe the local behaviour of the output at the specific input value, without considering how the output behaves over a range of inputs.

\subsection{Separating Positive and Negative Contributions}

We will see in \textbf{Section 3.5.3} that, in some situations, it is essential to treat positive and negative contributions differently. To do this, for every neuron $y$, we will introduce $\Delta y^+$ and $\Delta y^-$ to represent the positive and negative components of $\Delta y$, such that:
\begin{align*}
\Delta y &= \Delta y^+ + \Delta y^-\\
C_{\Delta y \Delta t} &= C_{\Delta y^+ \Delta t} + C_{\Delta y^- \Delta t} 
\end{align*}

For linear neurons, $\Delta y^+$ and $\Delta y^-$ are found by writing $\Delta y$ as a sum of terms involving its inputs $\Delta x_i$ and grouping positive and negative terms together. The importance of this will become apparent when applying the RevealCancel rule (\textbf{Section 3.5.3}), where for a given target neuron $t$ we may find that $m_{\Delta y^+ \Delta t}$ and $m_{\Delta y^- \Delta t}$ differ. However, when applying only the Linear or Rescale rules (\textbf{Section 3.5.1} and \textbf{Section 3.5.2}), $m_{\Delta y \Delta t} = m_{\Delta y^+ \Delta t} = m_{\Delta y^- \Delta t}$.

\subsection{Rules for Assigning Contribution Scores}

We present the rules for assigning contribution scores for each neuron to its immediate inputs. In conjunction with the chain rule for multipliers (\textbf{Section 3.2}), these rules can be used to find the contributions of any input (not just the immediate inputs) to a target output via backpropagation.

\subsubsection{The Linear Rule}

This applies to Dense and Convolutional layers (excluding nonlinearities). Let $y$ be a linear function of its inputs $x_i$ such that $y = b + \sum_i w_i x_i$. We have $\Delta y = \sum_i w_i \Delta x_i$. We define the positive and negative parts of $\Delta y$ as:
\begin{align*}
\Delta y^+ &= \sum_i 1\{w_i \Delta x_i > 0\} w_i\Delta x_i\\
           &= \sum_i 1\{w_i \Delta x_i > 0\} w_i (\Delta x_i^+ + \Delta x_i^-)\\
\Delta y^- &= \sum_i 1\{w_i \Delta x_i < 0\} w_i \Delta x_i\\
           &= \sum_i 1\{w_i \Delta x_i < 0\} w_i (\Delta x_i^+ + \Delta x_i^-)
\end{align*}

Which leads to the following choice for the contributions:\\
$C_{\Delta x_i^+ \Delta y^+} = 1\{w_i \Delta x_i > 0\} w_i \Delta x_i^+$\\
$C_{\Delta x_i^- \Delta y^+} = 1\{w_i \Delta x_i > 0\} w_i \Delta x_i^-$\\
$C_{\Delta x_i^+ \Delta y^-} = 1\{w_i \Delta x_i < 0\} w_i \Delta x_i^+$\\
$C_{\Delta x_i^- \Delta y^-} = 1\{w_i \Delta x_i < 0\} w_i \Delta x_i^-$\\

We can then find multipliers using the definition in \textbf{Section 3.2.1}, which gives $m_{\Delta x_i^+ \Delta y^+} = m_{\Delta x_i^- \Delta y^+} = 1\{w_i \Delta x_i > 0\} w_i$ and $m_{\Delta x_i^+ \Delta y^-} = m_{\Delta x_i^- \Delta y^-} = 1\{w_i \Delta x_i < 0\} w_i$.

What about when $\Delta x_i = 0$? While setting multipliers to 0 in this case would be consistent with summation-to-delta, it is possible that $\Delta x_i^+$ and $\Delta x_i^-$ are nonzero (and cancel each other out), in which case setting the multiplier to 0 would fail to propagate importance to them. To avoid this, we set $m_{\Delta x_i^+ \Delta y^+} = m_{\Delta x_i^+ \Delta y^-} = 0.5w_i$ when $\Delta x_i$ is 0 (similarly for $\Delta x^-$). See \textbf{Appendix B} for how to compute these multipliers using standard neural network ops.

\subsubsection{The Rescale Rule}

This rule applies to nonlinear transformations that take a single input, such as the ReLU, tanh or sigmoid operations. Let neuron $y$ be a nonlinear transformation of its input $x$ such that $y = f(x)$. Because $y$ has only one input, we have by summation-to-delta that $C_{\Delta x \Delta y} = \Delta y$, and consequently $m_{\Delta x \Delta y} = \frac{\Delta y}{\Delta x}$. For the Rescale rule, we set $\Delta y^+$ and $\Delta y^-$ proportional to $\Delta x^+$ and $\Delta x^-$ as follows:
\begin{align*}
\Delta y^+ = \frac{\Delta y}{\Delta x} \Delta x^+ = C_{\Delta x^+ \Delta y^+}\\
\Delta y^- = \frac{\Delta y}{\Delta x} \Delta x^- = C_{\Delta x^- \Delta y^-}
\end{align*}
Based on this, we get:
\begin{align*}
 m_{\Delta x^+ \Delta y^+} = m_{\Delta x^- \Delta y^-} = m_{\Delta x \Delta y} = \frac{\Delta y}{\Delta x}
\end{align*}
In the case where $x \rightarrow x^0$, we have $\Delta x \rightarrow 0$ and $\Delta y \rightarrow 0$. The definition of the multiplier approaches the derivative, i.e. $m_{\Delta x \Delta y} \rightarrow \frac{dy}{dx}$, where the $\frac{dy}{dx}$ is evaluated at $x = x^0$. We can thus use the gradient instead of the multiplier when $x$ is close to its reference to avoid numerical instability issues caused by having a small denominator.

Note that the Rescale rule addresses both the saturation and the thresholding problems illustrated in \textbf{Fig. 1} and \textbf{Fig. 2}. In the case of \textbf{Fig. 1}, if $i_1^0 = i_2^0 = 0$, then at $i_1 + i_2 > 1$ we have $\Delta h = -1$ and $\Delta y = 1$, giving $m_{\Delta h \Delta y} = \frac{\Delta y}{\Delta h} = -1$ even though $\frac{dy}{dh} = 0$ (in other words, using difference-from-reference allows information to flow even when the gradient is zero). In the case of \textbf{Fig. 2}, assuming $x^0 = y^0 = 0$, at $x = 10 + \epsilon$ we have $\Delta y = \epsilon$, giving $m_{\Delta x \Delta y} = \frac{\epsilon}{10 + \epsilon}$ and $C_{\Delta x \Delta y} = \Delta x \times m_{\Delta x \Delta y}  = \epsilon$. By contrast, gradient$\times$input assigns a contribution of $10 + \epsilon$ to $x$ and $-10$ to the bias term (DeepLIFT never assigns importance to bias terms).

As revealed in previous work \cite{LundbergL16}, there is a connection between DeepLIFT and Shapely values. Briefly, the Shapely values measure the average marginal effect of including an input over all possible orderings in which inputs can be included. If we define ``including" an input as setting it to its actual value instead of its reference value, DeepLIFT can be thought of as a fast approximation of the Shapely values. At the time, Lundberg \& Lee cited a preprint of DeepLIFT which described only the Linear and Rescale rules with no separate treatment of positive and negative contributions.

\subsubsection{An Improved Approximation of the Shapely Values: The RevealCancel Rule}

While the Rescale rule improves upon simply using gradients, there are still some situations where it can provide misleading results. Consider the $\min(i_1, i_2)$ operation depicted in \textbf{Fig. 3}, with reference values of $i_1 = 0$ and $i_2 = 0$. Using the Rescale rule, all importance would be assigned either to $i_1$ or to $i_2$ (whichever is smaller). This can obscure the fact that both inputs are relevant for the $\min$ operation.

To understand why this occurs, consider the case when $i_1 > i_2$. We have $h_1 = (i_1 - i_2) > 0$ and $h_2 = \max(0,h_1) = h_1$. By the Linear rule, we calculate that $C_{\Delta i_1 \Delta h_1} = i_1$ and $C_{\Delta i_2 \Delta h_1} = -i_2$. By the Rescale rule, the multiplier $m_{\Delta h_1 \Delta h_2}$ is $\frac{\Delta h_2}{\Delta h_1} = 1$, and thus $C_{\Delta i_1 \Delta h_2} = m_{\Delta h_1 \Delta h_2} C_{\Delta i_1 \Delta h_1} = i_1$ and $C_{\Delta i_2 \Delta h_2} = m_{\Delta h_1 \Delta h_2} C_{\Delta i_2 \Delta h_1} = -i_2$. The total contribution of $i_1$ to the output $o$ becomes $(i_1 - C_{\Delta i_1 \Delta h_2}) = (i_1 - i_1) = 0$, and the total contribution of $i_2$ to $o$ is $-C_{\Delta i_2 \Delta h_2} = i_2$. This calculation is misleading as it discounts the fact that $C_{\Delta i_2 \Delta h_2}$ would be $0$ if $i_1$ were 0 - in other words, it ignores a dependency induced between $i_1$ and $i_2$ that comes from $i_2$ canceling out $i_1$ in the nonlinear neuron $h_2$. A similar failure occurs when $i_1 < i_2$; the Rescale rule results in $C_{\Delta i_1 \Delta o} = i_1$ and $C_{\Delta i_2 \Delta o} = 0$. Note that gradients, gradient$\times$input, Guided Backpropagation and integrated gradients would also assign all importance to either $i_1$ or $i_2$, because for any given input the gradient is zero for one of $i_1$ or $i_2$ (see \textbf{Appendix C} for a detailed calculation).

One way to address this is by treating the positive and negative contributions separately. We again consider the nonlinear neuron $y = f(x)$. Instead of assuming that $\Delta y^+$ and $\Delta y^-$ are proportional to $\Delta x^+$ and $\Delta x^-$ and that $m_{\Delta x^+ \Delta y^+} = m_{\Delta x^- \Delta y^-} = m_{\Delta x \Delta y}$ (as is done for the Rescale rule), we define them as follows:
\vspace{-5pt}
\begin{align*}
\Delta y^+ &=  \frac{1}{2} \left( f(x^0 + \Delta x^+) - f(x^0) \right) \\
           &+ \frac{1}{2} \left( f(x^0 + \Delta x^- + \Delta x^+) - f(x^0 + \Delta x^-) \right)\\
\Delta y^- &=  \frac{1}{2} \left( f(x^0 + \Delta x^-) - f(x^0) \right) \\
           &+ \frac{1}{2} \left( f(x^0 + \Delta x^+ + \Delta x^-) - f(x^0 + \Delta x^+) \right)\\
m_{\Delta x^+ \Delta y^+} &= \frac{C_{\Delta x^+ y^+}}{\Delta x^+} = \frac{\Delta y^+}{\Delta x^+}\text{ ; } m_{\Delta x^- \Delta y^-} = \frac{\Delta y^-}{\Delta x^-}
\end{align*}

In other words, we set $\Delta y^+$ to the average impact of $\Delta x^+$ after no terms have been added and after $\Delta x^-$ has been added, and we set $\Delta y^-$ to the average impact of $\Delta x^-$ after no terms have been added and after $\Delta x^+$ has been added. This can be thought of as the Shapely values of $\Delta x^+$ and $\Delta x^-$ contributing to $y$.

By considering the impact of the positive terms in the absence of negative terms, and the impact of negative terms in the absence of positive terms, we alleviate some of the issues that arise from positive and negative terms canceling each other out. In the case of \textbf{Fig. 3}, RevealCancel would assign a contribution of $0.5 \min(i_1, i_2)$ to both inputs (see \textbf{Appendix C} for a detailed calculation).

While the RevealCancel rule also avoids the saturation and thresholding pitfalls illustrated in \textbf{Fig. 1} and \textbf{Fig. 2}, there are some circumstances where we might prefer to use the Rescale rule. Specifically, consider a thresholded ReLU where $\Delta y > 0 \text{ iff } \Delta x \ge b$. If $\Delta x < b$ merely indicates noise, we would want to assign contributions of 0 to both $\Delta x^+$ and $\Delta x^-$ (as done by the Rescale rule) to mitigate the noise. RevealCancel may assign nonzero contributions  by considering $\Delta x^+$ in the absence of $\Delta x^-$ and vice versa.

\begin{figure}[!htb]
\centering
\includegraphics[width=0.4\textwidth]{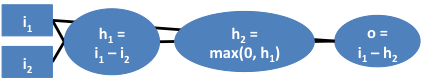}
\label{fig:min}
\caption{Network computing $o = \min(i_1, i_2)$. Assume $i_1^0 = i_2^0 = 0$. When $i_1 < i_2$ then $\frac{dy}{di_2}=0$, and when $i_2 < i_1$ then $\frac{do}{di_1} = 0$. Using any of the backpropagation approaches described in \textbf{Section 2.2} would result in importance assigned either exclusively to $i_1$ or $i_2$. With the RevealCancel rule, the net assigns $0.5\min(i_1, i_2)$ importance to both inputs.}
\end{figure}

\subsection{Choice of Target Layer}

In the case of softmax or sigmoid outputs, we may prefer to compute contributions to the linear layer preceding the final nonlinearity rather than the final nonlinearity itself. This would be to avoid an attentuation caused by the summation-to-delta property described in \textbf{Section 3.1}. For example, consider a sigmoid output $o = \sigma(y)$, where $y$ is the logit of the sigmoid function. Assume $y = x_1 + x_2$, where $x_1^0 = x_2^0 = 0$. When $x_1 = 50$ and $x_2 = 0$, the output $o$ saturates at very close to $1$ and the contributions of $x_1$ and $x_2$ are $0.5$ and $0$ respectively. However, when $x_1 = 100$ and $x_2 = 100$, the output $o$ is still very close to $1$, but the contributions of $x_1$ and $x_2$ are now both $0.25$. This can be misleading when comparing scores across different inputs because a stronger contribution to the logit would not always translate into a higher DeepLIFT score. To avoid this, we compute contributions to $y$ rather than $o$.

\textbf{Adjustments for Softmax Layers}

If we compute contributions to the linear layer preceding the softmax rather than the softmax output, an issue that could arise is that the final softmax output involves a normalization over all classes, but the linear layer before the softmax does not. To address this, we can normalize the contributions to the linear layer by subtracting the mean contribution to all classes. Formally, if $n$ is the number of classes, $C_{\Delta x \Delta c_i}$ represents the unnormalized contribution to class $c_i$ in the linear layer and $C'_{\Delta x \Delta c_i}$ represents the normalized contribution, we have:
\begin{equation}
C'_{\Delta x \Delta c_i} = C_{\Delta x \Delta c_i} - \frac{1}{n} \sum_{j=1}^{n} C_{\Delta x \Delta c_j}
\end{equation}
As a justification for this normalization, we note that subtracting a fixed value from all the inputs to the softmax leaves the output of the softmax unchanged.
%

\section{Results}

\subsection{Digit Classification (MNIST)}

We train a convolutional neural network on MNIST \cite{lecun1999} using Keras \cite{chollet2015} to perform digit classification and obtain 99.2\% test-set accuracy. The architecture consists of two convolutional layers, followed by a fully connected layer, followed by the softmax output layer (see \textbf{Appendix D} for full details on model architecture and training). We used convolutions with stride $>1$ instead of pooling layers, which did not result in a drop in performance as is consistent with previous work \cite{SpringenbergDBR14}. For DeepLIFT and integrated gradients, we used a reference input of all zeros.

To evaluate importance scores obtained by different methods, we design the following task: given an image that originally belongs to class $c_o$, we identify which pixels to erase to convert the image to some target class $c_t$. We do this by finding $S_{x_i\text{diff}} = S_{x_i c_o} - S_{x_i c_t}$ (where $S_{x_i c}$ is the score for pixel $x_i$ and class $c$) and erasing up to 157 pixels (20\% of the image) ranked in descending order of $S_{x_i\text{diff}}$ for which $S_{x_i\text{diff}} > 0$. We then evaluate the change in the log-odds score between classes $c_o$ and $c_t$ for the original image and the image with the pixels erased.

As shown in $\textbf{Fig. 4}$, DeepLIFT with the RevealCancel rule outperformed the other backpropagation-based methods. Integrated gradients (\textbf{Section 2.2.3}) computed numerically over either 5 or 10 intervals produced results comparable to each other, suggesting that adding more intervals would not change the result. Integrated gradients also performed comparably to gradient*input, suggesting that saturation and thresholding failure modes are not common on MNIST data. Guided Backprop discards negative gradients during backpropagation, perhaps explaining its poor performance at discriminating between classes. We also explored using the Rescale rule instead of RevealCancel on various layers and found that it degraded performance (\textbf{Appendix E}).

\begin{figure}[!bh]
\centering 
\includegraphics[width=0.49\textwidth, height=291pt]{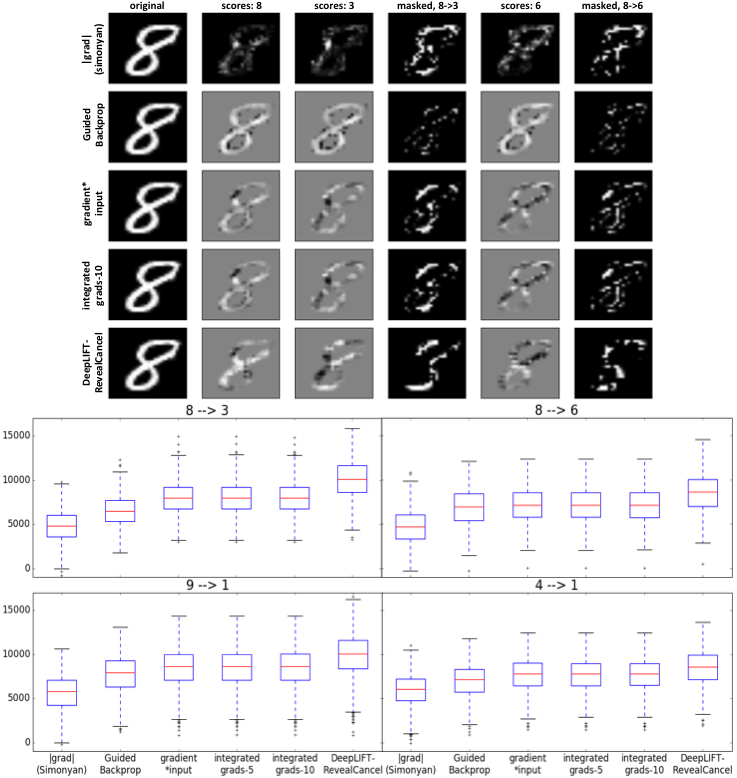}
\label{fig:MNIST_fig1}
\caption{\textbf{DeepLIFT with the RevealCancel rule better identifies pixels to convert one digit to another.} Top: result of masking pixels ranked as most important for the original class (8) relative to the target class (3 or 6). Importance scores for class 8, 3 and 6 are also shown. The selected image had the highest change in log-odds scores for the 8$\rightarrow$6 conversion using gradient*input or integrated gradients to rank pixels. Bottom: boxplots of increase in log-odds scores of target vs. original class after the mask is applied, for 1K images belonging to the original class in the testing set. ``Integrated gradients-n" refers to numerically integrating the gradients over $n$ evenly-spaced intervals using the midpoint rule.}

\end{figure}

\begin{figure*}[!htb]
\centering
\includegraphics[width=0.9\textwidth, height=290pt]{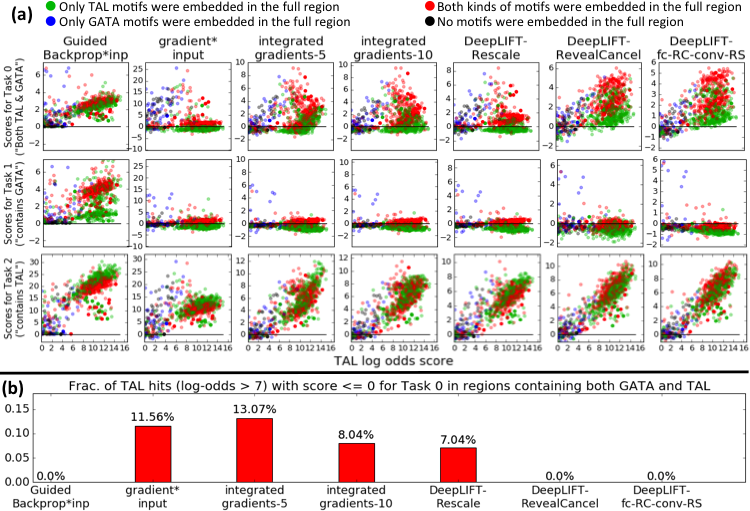}
\label{fig:TALGATA_talfig}
\caption{\textbf{DeepLIFT with RevealCancel gives qualitatively desirable behavior on TAL-GATA simulation}. (a) Scatter plots of importance score vs. strength of TAL1 motif match for different tasks and methods (see \textbf{Appendix G} for GATA1). For each region, top 5 motif matches are plotted. X-axes: log-odds of TAL1 motif match vs. background. Y-axes: total importance assigned to the match for specified task. Red dots are from regions where both TAL1 and GATA1 motifs were inserted during simulation; blue have GATA1 only, green have TAL1 only, black have no motifs inserted. ``DeepLIFT-fc-RC-conv-RS" refers to using RevealCancel on the fully-connected layer and Rescale on the convolutional layers, which appears to reduce noise relative to using RevealCancel on all layers. (b) proportion of strong matches (log-odds $>$ 7) to TAL1 motif in regions containing both TAL1 and GATA1 that had total score $\le$ 0 for task 0; Guided Backprop$\times$inp and DeepLIFT with RevealCancel have no false negatives, but Guided Backprop has false positives for Task 1 (Panel (a))}
\end{figure*}

\subsection{Classifying Regulatory DNA (Genomics)}

Next, we compared the importance scoring methods when applied to classification tasks on DNA sequence inputs (strings over the alphabet \{A,C,G,T\}). The human genome has millions of DNA sequence elements (~200-1000 in length) containing specific combinations of short functional words to which regulatory proteins (RPs) bind to regulate gene activity. Each RP (e.g. GATA1) has binding affinity to specific collections of short DNA words (motifs) (e.g. GATAA and GATTA).  A key problem in computational genomics is the discovery of motifs in regulatory DNA elements that give rise to distinct molecular signatures (labels) which can be measured experimentally. Here, in order to benchmark DeepLIFT and competing methods to uncover predictive patterns in DNA sequences, we design a simple simulation that captures the essence of the motif discovery problem described above. 

Background DNA sequences of length 200 were generated by sampling the letters ACGT at each position with probabilities $0.3, 0.2, 0.2$ and $0.3$ respectively. Motif instances were randomly sampled from previously known probabilistic motif models (See \textbf{Appendix F}) of two RPs named GATA1 and TAL1 (\textbf{Fig. 6a})\cite{kheradpour2014systematic}, and 0-3 instances of a given motif were inserted at random non-overlapping positions in the DNA sequences. We trained a multi-task neural network with two convolutional layers, global average pooling and one fully-connected layer on 3 binary classification tasks. Positive labeled sequences in task 1 represented ``both GATA1 and TAL1 present", task 2 represented ``GATA1 present" and in task 3 represented ``TAL1 present". $\frac{1}{4}$ of sequences had both GATA1 and TAL1 motifs (labeled 111), $\frac{1}{4}$ had only GATA1 (labeled 010), $\frac{1}{4}$ had only TAL1 (labeled 001), and $\frac{1}{4}$ had no motifs (labeled 000). Details of the simulation, network architecture and predictive performance are given in \textbf{Appendix F}. For DeepLIFT and integrated gradients, we used a reference input that had the expected frequencies of ACGT at each position (i.e. we set the ACGT channel axis to $0.3, 0.2, 0.2, 0.3$; see \textbf{Appendix J} for results using shuffled sequences as a reference). For fair comparison, this reference was also used for gradient$\times$input and Guided Backprop$\times$input (``input" is more accurately called $\Delta$input where $\Delta$ measured w.r.t the reference). For DNA sequence inputs, we found Guided Backprop$\times$input performed better than vanilla Guided Backprop; thus, we used the former.

Given a particular subsequence, it is possible to compute the log-odds score that the subsequence was sampled from a particular motif vs. originating from the background distribution of ACGT. To evaluate different importance-scoring methods, we found the top 5 matches (as ranked by their log-odds score) to each motif for each sequence from the test set, as well as the total importance allocated to the match by different importance-scoring methods for each task. The results are shown in \textbf{Fig. 5} (for TAL1) and \textbf{Appendix E} (for GATA1). Ideally, we expect an importance scoring method to show the following properties: (1) high scores for TAL1 motifs on task 2 and (2) low scores for TAL1 on task 1, with (3) higher scores corresponding to stronger log-odds matches; analogous pattern for GATA1 motifs (high for task 1, low for task 2); (4) high scores for both TAL1 and GATA1 motifs for task 0, with (5) higher scores on sequences containing both kinds of motifs vs. sequences containing only one kind (revealing cooperativity; corresponds to red dots lying above green dots in \textbf{Fig. 5}).

We observe Guided Backprop$\times$input fails (2) by assigning positive importance to TAL1 on task 1 (see \textbf{Appendix H} for an example sequence). It fails property (4) by failing to identify cooperativity in task 0 (red dots overlay green dots). Both Guided Backprop$\times$input and gradient$\times$input show suboptimal behavior regarding property (3), in that there is a sudden increase in importance when the log-odds score is around 7, but little differentiation at higher log-odds scores (by contrast, the other methods show a more gradual increase). As a result, Guided Backprop$\times$input and gradient$\times$input can assign unduly high importance to weak motif matches (\textbf{Fig. 6}). This is a practical consequence of the thresholding problem from \textbf{Fig. 2}. The large discontinuous jumps in gradient also result in inflated scores (note the scale on the y-axes) relative to other methods.

We explored three versions of DeepLIFT: Rescale at all nonlinearities (DeepLIFT-Rescale), RevealCancel at all nonlinearities (DeepLIFT-RevealCancel), and Rescale at convolutional layers with RevealCancel at the fully connected layer (DeepLIFT-fc-RC-conv-RS). In contrast to the results on MNIST, we found that DeepLIFT-fc-RC-conv-RS reduced noise relative to pure RevealCancel. We think this is because of the noise-suppression property discussed in \textbf{Section 3.5.3}; if the convolutional layers act like motif detectors, the input to convolutional neurons that do not fire may just represent noise and importance should not be propagated to them (see \textbf{Fig. 6} for an example sequence).

Gradient$\times$inp, integrated gradients and DeepLIFT-Rescale occasionally miss relevance of TAL1 for Task 0 (\textbf{Fig. 5b}), which is corrected by using RevealCancel on the fully connected layer (see example sequence in \textbf{Fig. 6}). Note that the RevealCancel scores seem to be tiered. As illustrated in \textbf{Appendix I}, this is related to having multiple instances of a given motif in a sequence (eg: when there are multiple TAL1 motifs, the importance assigned to the presence of TAL1 is distributed across all the motifs).

\begin{figure}[!htb]
\centering
\includegraphics[width=0.5\textwidth, height=140pt]{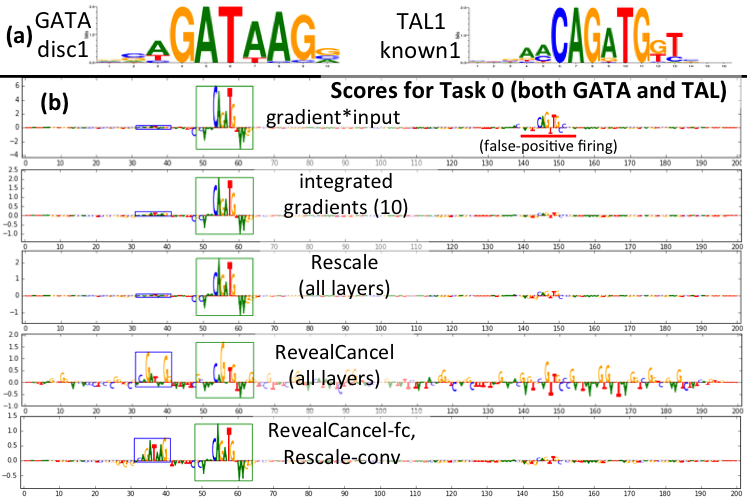}
\label{fig:TALGATA_fig3}
\caption{\textbf{RevealCancel highlights both TAL1 and GATA1 motifs for Task 0}. (a) PWM representations of the GATA1 motif and TAL1 motif used in the simulation (b) Scores for example sequence containing both TAL1 and GATA1 motifs. Letter height reflects the score. Blue box is location of embedded GATA1 motif, green box is location of embedded TAL1 motif. Red underline is chance occurrence of weak match to TAL1 (CAGTTG instead of CAGATG). Both TAL1 and GATA1 motifs should be highlighted for Task 0. RevealCancel on only the fully-connected layer reduces noise compared to RevealCancel on all layers.}
\end{figure}

\section{Conclusion}

We have presented DeepLIFT, a novel approach for computing importance scores based on explaining the difference of the output from some `reference' output in terms of differences of the inputs from their `reference' inputs. Using the difference-from-reference allows information to propagate even when the gradient is zero (\textbf{Fig. 1}), which could prove especially useful in Recurrent Neural Networks where saturating activations like sigmoid or tanh are popular. DeepLIFT avoids placing potentially misleading importance on bias terms (in contrast to gradient*input - see \textbf{Fig. 2}). By allowing separate treatment of positive and negative contributions, the DeepLIFT-RevealCancel rule can identify dependencies missed by other methods (\textbf{Fig. 3}). Open questions include how to apply DeepLIFT to RNNs, how to compute a good reference empirically from the data, and how best to propagate importance through `max' operations (as in Maxout or Maxpooling neurons) beyond simply using the gradients.

\newpage

\section{Appendix}
The appendix can be downloaded at:\\ \url{http://proceedings.mlr.press/v70/shrikumar17a/shrikumar17a-supp.pdf}

\bibliography{deeplift}

\begin{thebibliography}{14}
\providecommand{\natexlab}[1]{#1}
\providecommand{\url}[1]{\texttt{#1}}
\expandafter\ifx\csname urlstyle\endcsname\relax
  \providecommand{\doi}[1]{doi: #1}\else
  \providecommand{\doi}{doi: \begingroup \urlstyle{rm}\Url}\fi

\bibitem[Bach et~al.(2015)Bach, Binder, Montavon, Klauschen, M{\"{u}}ller, and
  Samek]{Bach2015-pq}
Bach, Sebastian, Binder, Alexander, Montavon, Gr\'{e}goire, Klauschen,
  Frederick, M{\"{u}}ller, Klaus-Robert, and Samek, Wojciech.
\newblock On {Pixel-Wise} explanations for {Non-Linear} classifier decisions by
  {Layer-Wise} relevance propagation.
\newblock \emph{PLoS One}, 10\penalty0 (7):\penalty0 e0130140, 10~July 2015.

\bibitem[Chollet(2015)]{chollet2015}
Chollet, François.
\newblock keras.
\newblock \url{https://github.com/fchollet/keras}, 2015.

\bibitem[Kheradpour \& Kellis(2014)Kheradpour and
  Kellis]{kheradpour2014systematic}
Kheradpour, Pouya and Kellis, Manolis.
\newblock Systematic discovery and characterization of regulatory motifs in
  encode tf binding experiments.
\newblock \emph{Nucleic acids research}, 42\penalty0 (5):\penalty0 2976--2987,
  2014.

\bibitem[Kindermans et~al.(2016)Kindermans, Schütt, Müller, and
  Dähne]{kindermans2016investigating}
Kindermans, Pieter-Jan, Schütt, Kristof, Müller, Klaus-Robert, and Dähne,
  Sven.
\newblock Investigating the influence of noise and distractors on the
  interpretation of neural networks.
\newblock \emph{CoRR}, abs/1611.07270, 2016.
\newblock URL \url{https://arxiv.org/abs/1611.07270}.

\bibitem[LeCun et~al.(1999)LeCun, Cortes, and Burges]{lecun1999}
LeCun, Yann, Cortes, Corinna, and Burges, Christopher~J.C.
\newblock The mnist database of handwritten digits.
\newblock \url{http://yann.lecun.com/exdb/mnist/}, 1999.

\bibitem[Lundberg \& Lee(2016)Lundberg and Lee]{LundbergL16}
Lundberg, Scott and Lee, Su{-}In.
\newblock An unexpected unity among methods for interpreting model predictions.
\newblock \emph{CoRR}, abs/1611.07478, 2016.
\newblock URL \url{http://arxiv.org/abs/1611.07478}.

\bibitem[Selvaraju et~al.(2016)Selvaraju, Das, Vedantam, Cogswell, Parikh, and
  Batra]{DBLP:journals/corr/SelvarajuDVCPB16}
Selvaraju, Ramprasaath~R., Das, Abhishek, Vedantam, Ramakrishna, Cogswell,
  Michael, Parikh, Devi, and Batra, Dhruv.
\newblock Grad-cam: Why did you say that? visual explanations from deep
  networks via gradient-based localization.
\newblock \emph{CoRR}, abs/1610.02391, 2016.
\newblock URL \url{http://arxiv.org/abs/1610.02391}.

\bibitem[Shrikumar et~al.(2016)Shrikumar, Greenside, Shcherbina, and
  Kundaje]{shrikumar2016not}
Shrikumar, Avanti, Greenside, Peyton, Shcherbina, Anna, and Kundaje, Anshul.
\newblock Not just a black box: Learning important features through propagating
  activation differences.
\newblock \emph{arXiv preprint arXiv:1605.01713}, 2016.

\bibitem[Simonyan et~al.(2013)Simonyan, Vedaldi, and
  Zisserman]{simonyan2013deep}
Simonyan, Karen, Vedaldi, Andrea, and Zisserman, Andrew.
\newblock Deep inside convolutional networks: Visualising image classification
  models and saliency maps.
\newblock \emph{arXiv preprint arXiv:1312.6034}, 2013.

\bibitem[Springenberg et~al.(2014)Springenberg, Dosovitskiy, Brox, and
  Riedmiller]{SpringenbergDBR14}
Springenberg, Jost~Tobias, Dosovitskiy, Alexey, Brox, Thomas, and Riedmiller,
  Martin~A.
\newblock Striving for simplicity: The all convolutional net.
\newblock \emph{CoRR}, abs/1412.6806, 2014.
\newblock URL \url{http://arxiv.org/abs/1412.6806}.

\bibitem[Sundararajan et~al.(2016)Sundararajan, Taly, and
  Yan]{SundararajanTY16}
Sundararajan, Mukund, Taly, Ankur, and Yan, Qiqi.
\newblock Gradients of counterfactuals.
\newblock \emph{CoRR}, abs/1611.02639, 2016.
\newblock URL \url{http://arxiv.org/abs/1611.02639}.

\bibitem[Zeiler \& Fergus(2013)Zeiler and Fergus]{zeiler2013visualizing}
Zeiler, Matthew~D. and Fergus, Rob.
\newblock Visualizing and understanding convolutional networks.
\newblock \emph{CoRR}, abs/1311.2901, 2013.
\newblock URL \url{http://arxiv.org/abs/1311.2901}.

\bibitem[Zhou \& Troyanskaya(2015)Zhou and Troyanskaya]{zhou2015predicting}
Zhou, Jian and Troyanskaya, Olga~G.
\newblock Predicting effects of noncoding variants with deep learning-based
  sequence model.
\newblock \emph{Nat Methods}, 12:\penalty0 931--4, 2015 Oct 2015.
\newblock ISSN 1548-7105.
\newblock \doi{10.1038/nmeth.3547}.

\bibitem[Zintgraf et~al.(2017)Zintgraf, Cohen, Adel, and
  Welling]{zintgraf2017Visualizing}
Zintgraf, Luisa~M, Cohen, Taco~S, Adel, Tameem, and Welling, Max.
\newblock Visualizing deep neural network decisions: Prediction difference
  analysis.
\newblock \emph{ICLR}, 2017.
\newblock URL \url{https://openreview.net/pdf?id=BJ5UeU9xx}.

\end{thebibliography}
\bibliographystyle{icml2017}

\section{Acknowledgements}
We thank Anna Shcherbina for early experiments applying DeepLIFT to image data and beta-testing. We thank Sinhan Kang of Korea University for identifying a typing error in Section 3.6.

\section{Funding}

AS was supported by a Howard Hughes Medical Institute International Student Research Fellowship and a Bio-X Bowes Fellowship. PG was supported by a Bio-X Stanford Interdisciplinary Graduate Fellowship. AK was supported by NIH grants DP2-GM-123485 and 1R01ES025009-02.

\section{Author Contributions}

AS \& PG conceptualized DeepLIFT. AS implemented DeepLIFT. AS ran experiments on MNIST. AS \& PG ran experiments on genomic data. AK provided guidance and feedback. AS, PG and AK wrote the manuscript.

\end{document}